\documentclass[letterpaper, 10pt, conference]{ieeeconf}

\IEEEoverridecommandlockouts
\usepackage{graphics}

\usepackage{amsmath}
\usepackage{amssymb}
\usepackage[colorlinks, linkcolor=blue, citecolor=blue]{hyperref}
\usepackage{cite}

\usepackage{algorithm}
\usepackage{url}
\usepackage{algpseudocode}
\usepackage{epsfig}
\usepackage{graphicx}
\usepackage{caption}
\usepackage{subfigure}
\usepackage{mathrsfs}
\usepackage{array}
\usepackage{multirow}
\usepackage{multicol}
\usepackage{booktabs}
\usepackage{makecell}
\usepackage[table]{xcolor}
\definecolor{revisionblue}{RGB}{0,0,180}
\newcommand{\rev}[1]{#1}

\usepackage{color}
\usepackage{amsfonts}
\usepackage{gensymb}
\usepackage{threeparttable}

\usepackage{amsthm}

\usepackage{tikz}
\usepackage{bbding}
\usetikzlibrary{positioning,arrows.meta,calc}

\title{\LARGE \bf
Learning Modular Policy for Multi-Floor Object Navigation:\\A Factorized Framework for Diagnostic Study
}

\author{
Shichao Zhai$^{1}$, Shuhao Ye$^{1}$, Rong Xiong$^{1}$, Yue Wang$^{1}$$^\dagger$  %
\thanks{$^{1}$Shichao Zhai, Shuhao Ye, Rong Xiong and Yue Wang are with the State Key Laboratory of Industrial Control and Technology, Zhejiang University, Hangzhou, China. Yue Wang is the corresponding author {\tt\footnotesize wangyue@iipc.zju.edu.cn}}%
}

\begin{document}

\maketitle
\thispagestyle{empty}
\pagestyle{empty}

\begin{abstract}
Object-goal navigation (ObjectNav) in multi-floor scenarios presents a challenge due to sparse rewards caused by long-horizon decision-making. In this paper, we propose a diagnostic study based on a modular framework with an effective learnable policy to analyze failure factors in multi-floor scenarios. To achieve an effective policy for diagnosis, we design the hierarchical factorization policy that deconstructs a single global policy into an intra-floor exploration policy and an inter-floor switching policy. To providing an effective initialization for Reinforcement Learning (RL), the lightweight intra-floor policy is learned by distilling the exploration logic of Visual Language Models (VLMs). 
Under idealized assumptions, we show that the factorized policy is theoretically equivalent to a single global policy at the policy-representation level. Experiment results indicate that perception performance and stair climbing stability are the primary bottlenecks in multi-floor navigation.

\end{abstract}

\section{Introduction}

Object-goal navigation (ObjectNav) requires an agent to reach an object goal in an unseen environment. Real-world scenarios often contain multiple floors, requiring the agent not only to make intra-floor decisions but also to determine when to switch floors. The main challenge lies in long-horizon decision-making, where the consequences of each decision may emerge only after many steps. Such long-horizon processes typically involve sparse rewards, making timely feedback difficult and ultimately causing leading to the failure of policy training.

The recent emergence of Vision-Language Models (VLMs) provides a promising paradigm for addressing these challenges. Without task-specific training, they can be directly used for decision-making, leading to the emergence of zero-shot methods with modular structures based on VLMs~\cite{gong2025stairway,zhang2024mfnp,zhang2025apexnav,yokoyama2024vlfm}. Meanwhile, learning-based end-to-end methods have achieved strong performance through large-scale imitation learning, reinforcement learning, or VLM fine-tuning~\cite{yokoyama2025filmnav,yadav2022ovrl,yadav2023ovrlv2,ramrakhya2023pirlnav,wasserman2023xgx}. Although modular designs improve interpretability and end-to-end methods achieve superior performance, the performance differences among existing methods do not directly reveal where the dominant bottlenecks arise in multi-floor scenarios. In particular, policy decision-making, perception, and low-level execution are coupled in the complete navigation process. Therefore, a diagnostic study with an effective framework is required to identify the main performance bottlenecks in modern multi-floor navigation methods.

To conduct a detailed diagnosis, an intuitive approach is to design a modular framework with an effective learnable policy, which can decouple performance bottlenecks induced by policy from other factors. However, due to the lack of supervised datasets for multi-floor navigation, it is impractical to directly train a single global policy through imitation learning (IL), and using online reinforcement learning (RL) to train a single global policy requires millions of environmental interactions, resulting in high time and computational costs. Given these limitations, exploiting the environment’s inherent structure to decompose a single global policy is a feasible approach for the learning of an effective policy.

\begin{figure}[t]
\centering
\includegraphics[scale=0.11]{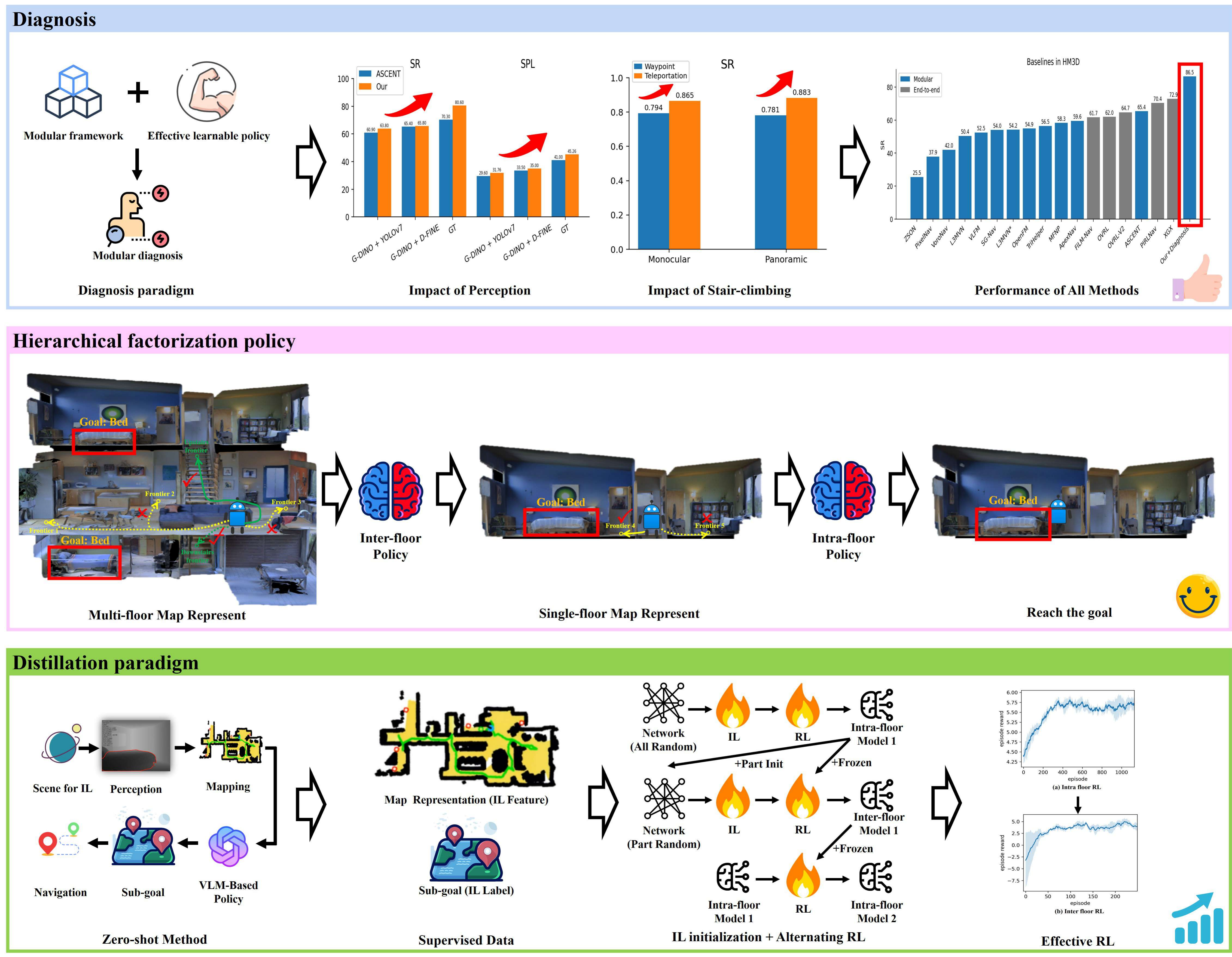}
\caption{\small{Description of diagnosis, factorization and distillation.}}
\label{First_img}
\vspace{-0.7cm}
\end{figure}

We address these constraints by designing the hierarchical factorization policy. Specifically, we decompose a single global policy into an intra-floor policy and an inter-floor policy based on the natural structure of the environment. To provide an effective initialization for subsequent RL, we utilize IL for knowledge distillation, transforming the exploration logic of VLMs into a lightweight intra-floor policy. 
After IL initialization, we use alternating RL as a practical optimization strategy, freezing one policy while updating the other to decouple floor-switching and sub-goal learning. Crucially, under idealized assumptions, we show that the factorized policy is theoretically equivalent to a single global policy at the policy-representation level.

Once the factorized policy converges, our modular framework enables systematic diagnosis of multi-floor navigation bottlenecks. The results show that the main limitations lie not in decision-making, but in perception and stair-climbing stability. With GT perception and ideal stair climbing, our framework outperforms existing methods. Our contributions are summarized as follows:

\begin{itemize}
    \item We provide a diagnostic study for multi-floor ObjectNav, which utilizes an effective learnable policy and a modular structure to decompose failure modes, indicating that perception reliability and stair climbing stability are two main bottlenecks.

    \item We design the hierarchical factorization policy that deconstructs the policy into an intra-floor policy and an inter-floor policy, %
    and show that, under idealized assumptions, the factorized policy is theoretically equivalent to a single global policy at the policy-representation level.

    \item We obtain a lightweight intra-floor policy by distilling the exploration logic of VLMs, enabling effective RL fine-tuning.
\end{itemize}

\section{Related Works}
\subsection{Modular Zero-shot Methods}
Previous studies have focused on solving intra-floor and inter-floor decision-making problems based on the zero-shot paradigm. For intra-floor navigation, SG-Nav~\cite{yin2024sgnav} prompts an LLM with an online-constructed 3D scene graph to reason about potential sub-goals. VLFM~\cite{yokoyama2024vlfm} uses a pre-trained VLM to produce a value map for scoring frontiers toward the object goal. ApexNav~\cite{zhang2025apexnav} further improves zero-shot exploration efficiency by adaptive switching between semantic-driven and geometry-driven exploration.

For inter-floor navigation, MFNP~\cite{zhang2024mfnp} extends zero-shot navigation to multi-floor environments with a rule-based inter-floor mechanism. ASCENT~\cite{gong2025stairway} emphasizes floor-aware exploration through hierarchical semantic mapping and LLM-driven coarse-to-fine frontier reasoning, enabling efficient multi-floor decision making without additional training. 

\subsection{End-to-End Methods}
End-to-end approaches directly map observations to actions using IL or RL. The OVRL series~\cite{yadav2022ovrl,yadav2023ovrlv2} demonstrates that visual representation can achieve robust ObjectNav performance without explicit modules. PIRLNav~\cite{ramrakhya2023pirlnav} pretrains policies with IL and further improves them through RL fine-tuning. XGX~\cite{wasserman2023xgx} proposes a learning-based exploration and exploitation coupling, where policies are jointly optimized to improve goal-conditioned exploration and approach behavior. Although end-to-end methods are simple and scalable, they lack explicit multi-floor representations and separate inter-floor and intra-floor decisions, making multi-floor decision-making less controllable.

Based on prior work, we diagnose the main bottlenecks using a
modular framework with a hierarchical factorization policy
consisting of inter-floor and intra-floor policies.

\begin{figure*}[ht]
\centering
\includegraphics[scale=0.5]
{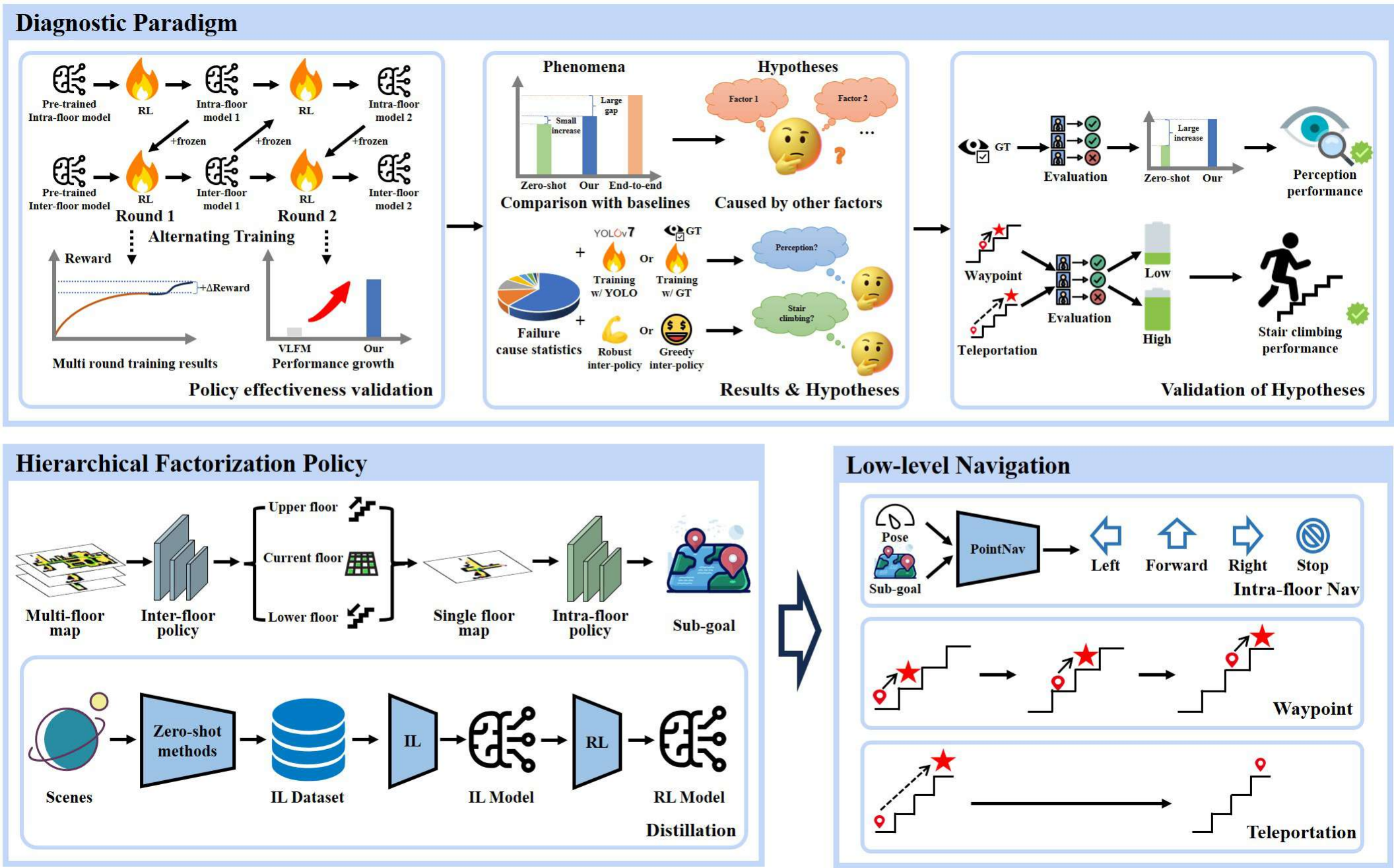}
\caption{\small{Overview. The Diagnostic Paradigm includes policy effectiveness validation, results \& hypotheses and validation of hypotheses to diagnose the causes of performance bottlenecks. The Hierarchical Factorization Policy module includes inter-floor policy and intra-floor policy. The Low-level Navigation module computes low-level actions for point-to-point navigation toward the sub-goal.}}
\label{Framework}
\vspace{-0.5cm}
\end{figure*}

\section{Task setup}

ObjectNav requires an agent to reach an object goal in an unseen environment. At each episode start, the agent is randomly initialized and assigned a goal category. At time $t$, the agent receives $O_t=(V_t,P_t)$, where $V_t$ is the RGBD image and $P_t$ is its pose, and selects from six actions: MoveForward, TurnLeft, TurnRight, LookUp, LookDown, and Stop. An episode succeeds if the agent stops within distance $d_{\rm succ}$ of the goal and is limited to $T_{\max}$ steps.

\section{Methodology}
\subsection{Overview}

Fig.~\ref{Framework} illustrates our framework, which uses a
diagnostic paradigm to identify the main bottlenecks in multi-floor
navigation. It employs a hierarchical factorization policy composed
of inter-floor and intra-floor policies, followed by the low-level
navigation policy described in Sec.~\ref{sec:low_level_nav}.

\subsection{Diagnostic Paradigm}

First, we validate policy effectiveness through multiple rounds of alternating intra-floor and inter-floor training. The policy is considered effective if each round improves performance and our method outperforms the zero-shot method.

Secondly, hypotheses will be proposed based on experimental results. If our method has limited improvement compared to the zero-shot multi-floor baselines and still has a significant gap compared to the end-to-end baselines, we believe that factors other than the policy cause the performance bottlenecks. Furthermore, combined with the failure cause statistics, if the success rate using ground-truth (GT) perception exceeds that of YOLO-based perception during training, then perception is hypothesized as one of the causes of performance bottlenecks. 
Similarly, if the learned and greedy inter-floor policies perform comparably under GT perception and waypoint mode, while both improve with teleportation, stair-climbing stability is hypothesized as another performance bottleneck.

Finally, we validate these hypotheses through a series of experiments. A clear advantage over zero-shot multi-floor baselines under GT perception identifies perception as a primary bottleneck, while gains from teleportation over waypoint mode indicate stair climbing as another bottleneck.

\subsection{Hierarchical Factorization Policy}
We design the hierarchical factorization policy with an intra-floor policy and an inter-floor policy. Below we describe map representation, intra-floor policy, inter-floor policy, network architecture and training paradigm.

\noindent\textbf{Map Representation.}
We maintain a local sliding-window multi-floor map representation $M_{multi}=\mathrm{concat}(M_l,M_c,M_u)$, where $M_l$, $M_c$, and $M_u$ correspond to floors $c_t-1$, $c_t$, and $c_t+1$, respectively. The window is re-centered after each floor transition, so the method is not limited to three-floor buildings. For map of each floor, $M_i$ is represented as a tensor with dimensions $6 \times W \times H$, where $6$, $W$ and $H$ denote the map’s channel, width and height, respectively. At time $t$, each channel of $M_i$ is represented as follows:  
\textbf{Semantic frontier map} $F^{t}$ is obtained by fusing all frontiers and all object-goal detections on the current floor. At time $t$, object-goal detection masks are projected onto the top-down map using depth and pose, and then fused with the previous semantic frontier map $F^{t-1}$ to obtain the updated map $F^t$. 
\textbf{Value map $U^t$} is obtained by masking the VLM-derived value map with the explored map and obstacle map, so that only explored and collision-free cells keep valid values, while obstacle cells are assigned $-1$ and the remaining invalid cells are assigned $0$. \textbf{Robot state map} $R^{t}$ is constructed by projecting the agent’s position at time $t$ onto the grid map and assigning its current heading angle to the corresponding location. The counterparts at time $t-1$ are $(F^{t-1}, U^{t-1})$. \textbf{Previous action map} $A^{t-1}$ is encoded by the selected sub-goal action at time $t-1$. Concretely, the single-floor map is given by
\begin{equation}
M_{i}=\mathrm{concat}\!\left(F^{t},\, U^{t},\, R^{t},\, F^{t-1},\, U^{t-1},\, A^{t-1}\right)
\end{equation}

When a staircase is detected, its mask is projected to the top-down map using depth and pose, and the nearest navigable cell around the projected staircase region is used as the staircase frontier. For the upper-floor representation $M_{u}$, when there exists a staircase frontier leading to the upper floor, we use the same construction as $M_{c}$ and the robot state map $R^{t}$ in $M_{u}$ is anchored at the upstair location with a fixed heading of $\pi$. Otherwise, we set $M_{u}$ to an all-zero tensor. The definition of lower-floor representation $M_{l}$ is similar.

\noindent\textbf{Intra-floor policy.}
Given the resized $\hat{M}_{c}\in\mathbb{R}^{6\times w\times h}$, the policy network outputs a logit map $Z^{t}\in\mathbb{R}^{w\times h}$. We further construct a binary mask map $\Omega^{t}\in\{0,1\}^{w\times h}$ from the resized $\hat{F}^t$ in $\hat{M}_c$ by marking the corresponding non-zero positions in $\hat{F}^{t}$ as 1 and all other locations as 0. The masked logits are computed as
\begin{equation}
\tilde{Z}^{t}_{x,y}=\Omega^{t}_{x,y}\,Z^{t}_{x,y}+(1-\Omega^{t}_{x,y})\cdot(-\infty),
\label{eq:mask_intra}
\end{equation}
which keeps the original logit values at valid candidate cells and assigns $-\infty$ to invalid cells. We then flatten $\tilde{Z}^{t}$ and apply a softmax to obtain a probability map $P^{t}$, and choose the intra-floor sub-goal as the argmax location on $P^{t}$. The intra-floor policy is conducted every $K_{\mathrm{intra}}$ simulator steps.

\noindent\textbf{Inter-floor policy.}
For floor switching, the policy takes the resized stacked map representation $\hat{M}_{multi}=\mathrm{concat}(\hat{M}_{l}, \hat{M}_{c}, \hat{M}_{u})$ as input and outputs a 3-dimensional logit vector $g^{t}\in\mathbb{R}^{3}$ corresponding to the actions \{Down, Stay, Up\}. We build a multi-floor policy mask $m^{t}\in\{0,1\}^{3}$ according to map validity. Specifically, if the resized $\hat{F}^t$ in $\hat{M}_c$ is an all-zero matrix, we set the mask entry of Stay to 0; otherwise it is 1. For the lower/upper floors, if either the resized $\hat{R}^t$ in $\hat{M}_u$/$\hat{M}_l$ is an all-zero matrix or the resized $\hat{F}^t$ in $\hat{M}_u$/$\hat{M}_l$ is an all-zero matrix, we set the corresponding mask entry to 0; otherwise it is 1. The masked logits are computed by
\begin{equation}
\tilde{g}^{t}_{j}=m^{t}_{j}\,g^{t}_{j}+(1-m^{t}_{j})\cdot(-\infty),\quad j\in\{\mathrm{Down},\mathrm{Stay},\mathrm{Up}\},
\label{eq:mask_inter}
\end{equation}
which is followed by a softmax to obtain a probability vector. The inter-floor decision result is generated by the argmax index and is executed every $K_{\mathrm{inter}}$ simulator steps. \rev{Therefore, a non-adjacent target floor can be reached by repeatedly applying adjacent floor-switching decisions as the local floor window is updated.}

\noindent\textbf{Network architecture.}
For intra-floor policy, a CNN encoder extracts and fuses
channel-wise features from $\hat{M}_c$, followed by a U-Net that
outputs $Z^{t}$. For the inter-floor policy, part of the intra-floor
network is reused to separately process resized $\hat{M}_{l}$,
$\hat{M}_{c}$, and $\hat{M}_{u}$. The resulting floor-wise features
are concatenated and passed through additional CNN layers and an MLP
head to produce the 3-dimensional logit vector $g^{t}$.

\noindent\textbf{Training paradigm.}
In the first stage, both policies are pretrained with IL. For the
intra-floor policy, VLFM~\cite{yokoyama2024vlfm} and an improved
variant collect supervised data, using resized $\hat{M}_c$ as inputs
and selected sub-goals as labels. For the inter-floor policy, a
greedy rule switches floors only after fully exploring the current
floor, with resized $(\hat{M}_l,\hat{M}_c,\hat{M}_u)$ as inputs and
the goal's relative floor (lower/current/upper) as labels. \rev{In the second stage, we use alternating RL as a practical optimization strategy rather than a theoretically guaranteed substitute for joint optimization. Since the two policies are executed at different frequencies, we freeze one policy and fine-tune the other in each round to reduce the optimization coupling between floor switching and sub-goal selection.}  We use Soft Actor-Critic (SAC) to perform RL fine-tuning of the intra-floor and inter-floor policies in each training round. The reward is defined as follows
\begin{equation}
r = 
\begin{cases} 
2.5, & \text{if success} \\
-(d^{\mathrm{now}}_{\mathrm{geo}}-d^{\mathrm{last}}_{\mathrm{geo}})-0.01\times\,\Delta k, & \text{otherwise}
\end{cases}
\end{equation}
\rev{where $d_{\rm geo}^{\rm last}$ and $d_{\rm geo}^{\rm now}$ are the geodesic distances before and after executing the selected sub-goal, and $\Delta k$ is the number of simulator steps between two consecutive decisions.}

\subsection{Low-level Navigation}\label{sec:low_level_nav}

Given a sub-goal, the low-level policy generates navigation actions.
A pretrained PointNav controller handles intra-floor navigation and
guides the agent to the staircase frontier for inter-floor navigation.
Stair traversal uses either teleportation, which resets the agent to
the target floor and adds 20 episode steps, or waypoint mode, which
guides PointNav across the staircase through successive waypoints.

\subsection{Equivalence of Factorized Policy and Global Policy}\label{sec:equivalence}

\noindent\textbf{Setup.}
We formulate multi-floor ObjectNav as a Markov Decision Process
$\mathcal{M}=(\mathcal{S},\mathcal{A}_t,P,r,\gamma)$. At time $t$,
$c_t$ denotes the agent's current floor, and $\mathcal{G}_t^{k}$
denotes the available sub-goals on floor $k$, including frontiers
and semantic goal positions. The global action space is

\begin{equation}
\mathcal{A}_t
\triangleq
\mathcal{G}_t^{c_t-1}\ \cup\ \mathcal{G}_t^{c_t}\ \cup\ \mathcal{G}_t^{c_t+1}.
\label{eq:joint_global_action_space}
\end{equation}
Each candidate belongs to a unique floor, so the sets
$\mathcal{G}_t^{c_t-1}, \mathcal{G}_t^{c_t}, \mathcal{G}_t^{c_t+1}$ are disjoint.

\noindent\textbf{Global policy.}
The single global policy $\pi_{\mathrm{G}}(a\mid s_t)$ directly selects a sub-goal
$a\in\mathcal{A}_t$.

\noindent\textbf{Factorized policy.}
The factorized policy comprises an inter-floor policy
$\pi_{\mathrm{inter}}(z\mid s_t)$ over
$z\in\{-1,0,+1\}$ and an intra-floor policy
$\pi_{\mathrm{intra}}(a\mid s_t,z)$ that selects
$a\in\mathcal{G}_t^{c_t+z}$ on the chosen floor. Define the expanded action space
\begin{equation}
\tilde{\mathcal{A}}_t
\triangleq
\{(z,a)\mid z\in\{-1,0,+1\},\ a\in\mathcal{G}_t^{c_t+z}\}.
\label{eq:joint_expanded_action_space}
\end{equation}
The factorized policy over $\tilde{\mathcal{A}}_t$ is
\begin{equation}
\pi_{\mathrm{Fac}}(z,a\mid s_t)
\triangleq
\pi_{\mathrm{inter}}(z\mid s_t)\,\pi_{\mathrm{intra}}(a\mid s_t,z).
\label{eq:joint_fac_policy}
\end{equation}
Since each global action
$a\in\mathcal{A}_t$ corresponds to a unique floor index $z(a)$. 
Therefore, the marginal action distribution induced by the factorized policy is
\begin{equation}
\begin{aligned}
\pi_{\mathrm{Fac}}(a\mid s_t) &= \sum_{z\in\{-1,0,+1\}} \pi_{\mathrm{Fac}}(z,a\mid s_t) \\
&= \pi_{\mathrm{inter}}(z(a)\mid s_t)\,\pi_{\mathrm{intra}}(a\mid s_t,z(a)).
\end{aligned}
\label{eq:joint_fac_marginal}
\end{equation}

\noindent\textbf{Proposition.}
Consider the maximum-entropy RL objective used by SAC:
\begin{equation}
J_{\mathrm{SAC}}(\pi)
\triangleq
\mathbb{E}_{\tau\sim\pi}
\Big[
\sum_{t\ge 0}\gamma^t\big(r(s_t,a_t)+\alpha\,\mathcal{H}(\pi(\cdot\mid s_t))\big)
\Big],
\label{eq:joint_sac_objective}
\end{equation}
where
$\mathcal{H}(\pi(\cdot\mid s))
=
-\mathbb{E}_{a\sim\pi(\cdot\mid s)}[\log \pi(a\mid s)]$.

Then optimizing SAC with a single global policy $\pi_{\mathrm{G}}(a\mid s)$ over $\mathcal{A}_t$
is equivalent to jointly optimizing SAC with a factorized policy $(\pi_{\mathrm{inter}},\pi_{\mathrm{intra}})$ over $\tilde{\mathcal{A}}_t$:
\begin{equation}
\max_{\pi_{\mathrm{G}}} J_{\mathrm{SAC}}(\pi_{\mathrm{G}})
=
\max_{\pi_{\mathrm{inter}},\,\pi_{\mathrm{intra}}} J_{\mathrm{SAC}}(\pi_{\mathrm{Fac}}).
\label{eq:joint_objective_equivalence}
\end{equation}
Thus, the optimal jointly trained factorized policy and the optimal directly trained global policy
achieve the same optimal value and induce the same optimal action distribution.

\noindent\textbf{Proof.}
We prove the equivalence by showing policy equality, reward equality, and entropy equality.

$\bullet$ \textbf{Policy equality.}
Given any global policy $\pi_{\mathrm{G}}(a\mid s)$ on $\mathcal{A}_t$, define
\begin{equation}
\begin{gathered}
\pi_{\mathrm{inter}}(z\mid s) = \sum_{a\in\mathcal{G}^{c+z}} \pi_{\mathrm{G}}(a\mid s), \\
\pi_{\mathrm{intra}}(a\mid s,z) = \frac{\pi_{\mathrm{G}}(a\mid s)}{\pi_{\mathrm{inter}}(z\mid s)}, \quad a\in\mathcal{G}^{c+z},
\end{gathered}
\label{eq:joint_construct_factorized}
\end{equation}
Then by \eqref{eq:joint_fac_marginal},
\begin{equation}
\pi_{\mathrm{Fac}}(a\mid s)
=
\pi_{\mathrm{inter}}(z(a)\mid s)\,\pi_{\mathrm{intra}}(a\mid s,z(a))
=
\pi_{\mathrm{G}}(a\mid s).
\label{eq:joint_same_marginal_1}
\end{equation}
Conversely, given any factorized pair $(\pi_{\mathrm{inter}},\pi_{\mathrm{intra}})$, define
\begin{equation}
\pi_{\mathrm{G}}(a\mid s)
=
\pi_{\mathrm{inter}}(z(a)\mid s)\,\pi_{\mathrm{intra}}(a\mid s,z(a)).
\label{eq:joint_construct_global}
\end{equation}
Therefore, the global policy $\pi_{\mathrm{G}}(a\mid s)$ and the factorized policy $\pi_{\mathrm{Fac}}(a\mid s)$ represent the same action distributions.

$\bullet$ \textbf{Reward equality.}
As $\pi_{\mathrm{G}}(a\mid s)=\pi_{\mathrm{Fac}}(a\mid s)$ for every state-action pair,
both policies induce the same distribution over trajectories $(s_t,a_t)$.
Since the transition probability $P$ and reward function $r$ depend only on the executed action $a_t$,
the expected discounted reward terms are equal:
\begin{equation}
\mathbb{E}_{\tau\sim\pi_{\mathrm{G}}}
\Big[\sum_{t\ge0}\gamma^t r(s_t,a_t)\Big]
=
\mathbb{E}_{\tau\sim\pi_{\mathrm{Fac}}}
\Big[\sum_{t\ge0}\gamma^t r(s_t,a_t)\Big].
\label{eq:joint_reward_equality}
\end{equation}

$\bullet$ \textbf{Entropy equality.}
Each action $a\in\mathcal{A}_t$ uniquely corresponds to $(z(a),a)\in\tilde{\mathcal{A}}_t$. Thus, for any state $s$,

\begin{equation}
\begin{aligned}
&\mathcal{H}(\pi_{\mathrm{G}}(\cdot\mid s)) = -\mathbb{E}_{a\sim\pi_{\mathrm{G}}}[\log \pi_{\mathrm{G}}(a\mid s)] \\
&= -\mathbb{E}_{(z,a)\sim\pi_{\mathrm{Fac}}}[\log \pi_{\mathrm{Fac}}(z,a\mid s)] = \mathcal{H}(\pi_{\mathrm{Fac}}(\cdot\mid s)).
\end{aligned}
\label{eq:joint_entropy_equality}
\end{equation}
Moreover, under the factorized form in \eqref{eq:joint_fac_policy},
\begin{equation}
\log \pi_{\mathrm{Fac}}(z,a\mid s)
=
\log \pi_{\mathrm{inter}}(z\mid s)
+
\log \pi_{\mathrm{intra}}(a\mid s,z),
\label{eq:joint_logprob_factorized}
\end{equation}
as used in joint SAC optimization of the factorized policy.

\rev{Combining policy equality, reward equality, and entropy equality, we obtain \eqref{eq:joint_objective_equivalence}. Therefore, under idealized assumptions, the factorized policy has the same representational capacity as a single global policy. 
\hfill $\blacksquare$}

\section{Experiments}

\subsection{Dataset}

We conduct experiments on HM3D~\cite{ramakrishnan2021hm3d} and
MP3D~\cite{chang2017matterport3d}. For the intra-floor
policy, IL uses 80 HM3D v0.1 scenes and RL fine-tuning uses 65
other HM3D v0.2 scenes. For the inter-floor policy, IL and RL use
25 and 23 multi-floor HM3D v0.1 scenes, respectively. Diagnostics
use the 20-scene HM3D v0.1 and 11-scene MP3D validation splits.

\subsection{Experiment Details}

\rev{We use D-FINE~\cite{peng2024dfine}, Grounding
DINO~\cite{liu2023groundingdino}, and
MobileSAM~\cite{zhang2023fsa} for perception, and
BLIP-2~\cite{li2023blip2} as the VLM. By default, all experiments use ground-truth (GT) perception for policy training and D-FINE-based perception for evaluation. Non-default settings, such as
YOLOv7 during RL training or GT perception during evaluation, are
reported in the corresponding tables. Maps are maintained at
$1000\times1000$ during map construction and resized to $500\times500$ before policy input.
The $\Delta k$ is set to 5 steps in intra-floor policy and is set to 50 steps in inter-floor policy.  We use the evaluation metrics Success Rate (SR) and Success weighted by Path Length (SPL)~\cite{batra2020objectnav}.} \rev{Runtime is measured in Habitat on an NVIDIA RTX 4090 GPU.
The average per-call runtimes are 98.76 ms for detection, 94.88 ms
for segmentation, 122.29 ms for VLM reasoning, 62.89 ms for
mapping, 3.29 ms for PointNav, 6.31 ms for the intra-floor policy,
12.55 ms for the inter-floor policy, and 10.07 ms for other
overhead. With the intra-floor and inter-floor policies invoked every
5 and 50 steps, respectively, the effective runtime is 393.69 ms
per step, corresponding to 2.54 FPS, with a peak GPU memory usage of 7899.7 MB.}

\begin{figure}[t]
\centering
\includegraphics[scale=0.28]{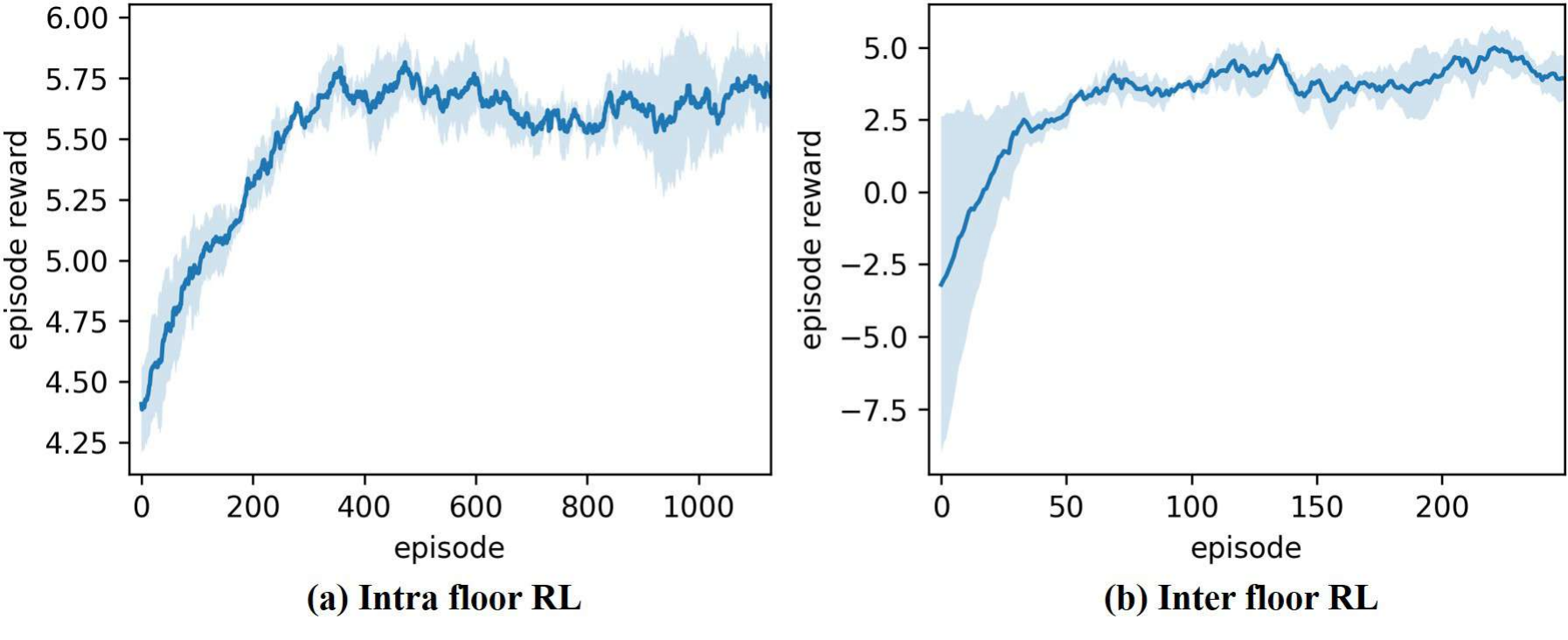}
\caption{\small{The trends of episode rewards for the first round of intra-floor policy and the inter-floor policy training. 
}}
\label{RL_Train_process}
\vspace{-0.2cm}
\end{figure}

\begin{figure}[t]
\centering
\includegraphics[scale=0.8]
{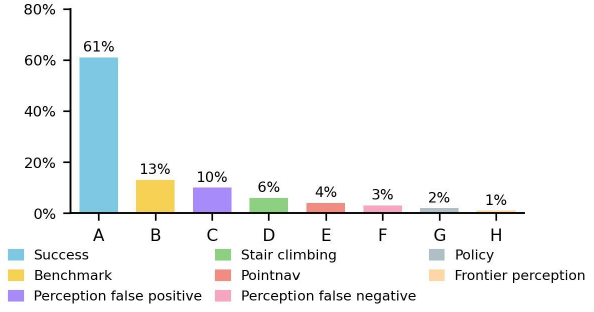}
\caption{\small{Failure cause statistics of our method.}}
\label{fig:failure_bar}
\vspace{-0.7cm}
\end{figure}

\subsection{Baselines}

We categorize baselines by four criteria: method type
(\textbf{modular} or \textbf{E2E}), multi-floor capability
(\textbf{single-floor} or \textbf{multi-floor}), learning requirement
in Habitat (\textbf{learning-free} or \textbf{learning-based}), and
success threshold (\textbf{0.1\,m}, \textbf{0.2\,m}, \textbf{1\,m},
or \textbf{unknown}).

\begin{table}[t]
\centering
\vspace{0.55cm}
\caption{\small{
Training performance of intra-floor policy in different rounds of training.
}}
\setlength{\tabcolsep}{12pt}
\renewcommand{\arraystretch}{0.2}
\scalebox{1.0}{
\begin{tabular}{cccc}
\toprule
\multicolumn{1}{c}{\textbf{Method}} & \multicolumn{1}{c}{\textbf{Policy}} & \multicolumn{1}{c}{\textbf{Stage}} & \multicolumn{1}{c}{\textbf{SR$\uparrow$}} \\\\
\midrule
Our & intra-floor & Round 1 & 87.06  \\\\
Our & intra-floor & Round 2 & \textbf{88.01} \\\\
\bottomrule
\end{tabular}
}
\label{tab:diff_round}
\vspace{-0.70cm}
\end{table}

\begin{table}[t]
\centering
\vspace{0.55cm}
\caption{\small{
Evaluation performance of our intra-floor policy and zero-shot data collector on HM3D.
}}
\setlength{\tabcolsep}{8pt}
\renewcommand{\arraystretch}{0.2}
\scalebox{1.0}{
\begin{tabular}{ccccc}
\toprule
\multicolumn{1}{c}{\textbf{Method}} & \multicolumn{1}{c}{\textbf{Learning}} & \multicolumn{1}{c}{\textbf{Policy}} & \multicolumn{1}{c}{\textbf{SR$\uparrow$}} & \multicolumn{1}{c}{\textbf{SPL$\uparrow$}} \\\\
\midrule
VLFM & zero-shot & -- & 52.5 & \textbf{30.4} \\\\
Our & IL & intra-floor & \underline{53.8} & 28.7 \\\\
Our & IL+RL (Round 1) & intra-floor & \textbf{57.5} & \underline{30.3} \\\\
\bottomrule
\end{tabular}
}
\label{tab:nav_performance_distill}
\vspace{-0.70cm}
\end{table}

\begin{table}[t]
\centering
\vspace{0.55cm}
\caption{\small{Evaluation performance of our model and baselines.}}
\label{tab:nav_performance_all}
\scalebox{1.0}{%
\setlength{\tabcolsep}{1.3pt}
\renewcommand{\arraystretch}{0.1}
\begin{tabular}{lcccccccc}
\toprule
\multirow{2}{*}{\textbf{Method}} & \multirow{2}{*}{\textbf{Type}} & \multirow{2}{*}{\textbf{Floor}} & \multirow{2}{*}{\textbf{Learning}} & 
\multirow{2}{*}[6.0pt]{\makecell[c]{\textbf{Success} \\ \textbf{thr.}}}
& \multicolumn{2}{c}{\textbf{HM3D}} & \multicolumn{2}{c}{\textbf{MP3D}}  \\
\cmidrule(lr){6-7}\cmidrule(lr){8-9}
& & & & & \textbf{SR$\uparrow$} & \textbf{SPL$\uparrow$} & \textbf{SR$\uparrow$} & \textbf{SPL$\uparrow$} \\
\midrule
VoroNav~\cite{wu2024voronav} & modular & single & \XSolidBrush & 0.1\,m & 42.0 & 26.0 & -- & -- \\
L3MVN~\cite{yu2023l3mvn} & modular & single & \XSolidBrush & 0.1\,m & 50.4 & 23.1 & 34.9 & 14.5 \\
VLFM~\cite{yokoyama2024vlfm} & modular & single & \XSolidBrush & 1\,m & 52.5 & 30.4 & 36.4 & 17.5 \\
SG-Nav~\cite{yin2024sgnav} & modular & single & \XSolidBrush & -- & 54.0 & 24.9 & \underline{40.2} & 16.0 \\
OpenFM~\cite{kuang2024openfmnav} & modular & single & \XSolidBrush & -- & 54.9 & 24.4 & 37.2 & 15.7 \\
TriHelper~\cite{zhang2024trihelper} & modular & single & \XSolidBrush & 0.1\,m & 56.5 & 25.3 & -- & -- \\
ApexNav~\cite{zhang2025apexnav} & modular & single & \XSolidBrush & 0.2\,m & \underline{59.6} &  \underline{33.0} & 39.2 & \underline{17.8} \\
\midrule
ZSON~\cite{majumdar2022zson} & modular & single & \Checkmark & -- & 25.5 & 12.6 & 15.3 & 4.8 \\
PixelNav~\cite{cai2023pixelnav} & modular & single & \Checkmark & -- & 37.9 & 20.5 & -- & -- \\
L3MVN*~\cite{yu2023l3mvn} & modular & single & \Checkmark & 0.1\,m & \underline{54.2} & \underline{23.1} & -- & -- \\
\midrule
MFNP~\cite{zhang2024mfnp} & modular & multi & \XSolidBrush & 0.1\,m & 58.3 & 26.7 & 41.1 & 15.4 \\
ASCENT~\cite{gong2025stairway} & modular & multi & \XSolidBrush & 1\,m & \underline{65.4} & \underline{33.5} & \textbf{44.5} & \underline{15.5} \\

\midrule

Our & modular & multi & \Checkmark & 0.1\,m & 63.7 & 33.9 & 40.0 & 17.2 \\
\rev{Our} & \rev{modular} & \rev{multi} & \rev{\Checkmark} & \rev{0.2\,m} & \rev{63.8} & \rev{34.0} & \rev{-} & \rev{-} 
\\
Our & modular & multi & \Checkmark & 1 \,m & 65.8 & 
\textbf{35.0} & \underline{42.6} & \textbf{18.2} \\
\midrule
\textcolor{gray}{FiLM-Nav~\cite{yokoyama2025filmnav}} & \textcolor{gray}{E2E} & \textcolor{gray}{single} & \textcolor{gray}{\Checkmark} & \textcolor{gray}{1\,m} & \textcolor{gray}{61.7} & \textcolor{gray}{\underline{37.3}} & \textcolor{gray}{--} & \textcolor{gray}{--} \\
\textcolor{gray}{OVRL~\cite{yadav2022ovrl}} & \textcolor{gray}{E2E} & \textcolor{gray}{multi} & \textcolor{gray}{\Checkmark} & \textcolor{gray}{0.1\,m} & \textcolor{gray}{62.0} & \textcolor{gray}{26.8} & \textcolor{gray}{--} & \textcolor{gray}{--} \\
\textcolor{gray}{OVRL-V2~\cite{yadav2023ovrlv2}} & \textcolor{gray}{E2E} & \textcolor{gray}{multi} & \textcolor{gray}{\Checkmark} & \textcolor{gray}{0.1\,m} & \textcolor{gray}{64.7} & \textcolor{gray}{28.1} & \textcolor{gray}{--} & \textcolor{gray}{--} \\
\textcolor{gray}{PIRLNav~\cite{ramrakhya2023pirlnav}} & \textcolor{gray}{E2E} & \textcolor{gray}{multi} & \textcolor{gray}{\Checkmark} & \textcolor{gray}{1\,m} & \textcolor{gray}{70.4} & \textcolor{gray}{34.1} & \textcolor{gray}{--} & \textcolor{gray}{--} \\
\textcolor{gray}{XGX~\cite{wasserman2023xgx}} & \textcolor{gray}{E2E} & \textcolor{gray}{multi} & \textcolor{gray}{\Checkmark} & \textcolor{gray}{1\,m} & \textcolor{gray}{\underline{72.9}} & \textcolor{gray}{35.7} & \textcolor{gray}{13.6} & \textcolor{gray}{5.1} \\
\bottomrule
\end{tabular}}
\vspace{-0.7cm}
\end{table}

\begin{table}[t]
\centering
\vspace{0.55cm}
\caption{\small{
\rev{Training performance of the intra-floor policy under different training-time perception settings.}
}}
\setlength{\tabcolsep}{13pt}
\renewcommand{\arraystretch}{0.3}
\scalebox{0.95}{
\begin{tabular}{ccccc}
\toprule
\multicolumn{1}{c}{\textbf{Policy}} & \multicolumn{1}{c}{\textbf{\rev{Training-time Perception}}} & \multicolumn{1}{c}{\textbf{SR$\uparrow$}} & \multicolumn{1}{c}{\textbf{SPL$\uparrow$}} \\\\
\midrule
intra-floor & YOLO-v7  & 72.9 & 40.0 \\\\
intra-floor & GT & \textbf{94.2} & \textbf{64.6} \\\\
\bottomrule
\end{tabular}
}
\label{tab:intrafloor_training_perform}
\vspace{-0.7cm}
\end{table}

\begin{table}[t]
\centering
\vspace{0.55cm}
\caption{\small{
Evaluation performance of different inter-floor policies under waypoint mode on HM3D.
}}
\setlength{\tabcolsep}{4pt}
\renewcommand{\arraystretch}{0.3}
\scalebox{1.05}{
\begin{tabular}{cccccc}
\toprule
\multicolumn{1}{c}{\textbf{Inter-floor policy}} & \multicolumn{1}{c}{\textbf{Perception}} & \multicolumn{1}{c}{\textbf{Stair climbing}} & \multicolumn{1}{c}{\textbf{SR$\uparrow$}} & \multicolumn{1}{c}{\textbf{SPL$\uparrow$}} \\\\
\midrule
Greedy & GT & Waypoint &\textbf{80.1} & \textbf{44.99} \\\\
IL+RL & GT & Waypoint & 79.4 & 44.61\\\\
\bottomrule
\end{tabular}
}
\label{tab:interfloor_ablation_mono}
\vspace{-0.7cm}
\end{table}

\begin{table}[t]
\centering
\vspace{0.45cm}
\caption{\small{
\rev{Evaluation performance of our model and modular baselines under test-time GT perception in HM3D.}
}}
\setlength{\tabcolsep}{9pt}
\renewcommand{\arraystretch}{0.2}
\scalebox{1.0}{
\begin{tabular}{ccccc}
\toprule
\multicolumn{1}{c}{\textbf{Method}} & \multicolumn{1}{c}{\textbf{Perception}} & \multicolumn{1}{c}{\textbf{Success thr.}} & \multicolumn{1}{c}{\textbf{SR$\uparrow$}} & \multicolumn{1}{c}{\textbf{SPL$\uparrow$}} \\\\
\midrule
VoroNav & GT & 0.1m & 67.6 & 40.5 \\\\
L3MVN & GT & 0.1m & 66.4 & 36.4 \\\\
ASCENT & GT & 1m & 70.3 & 41.0 \\\\
Our & GT & 0.1m & 79.3 & 44.5 \\\\
Our & GT & 1m & \textbf{80.6} & \textbf{45.3} \\\\
\bottomrule
\end{tabular}
}
\label{tab:nav_performance_all_gt}
\vspace{-0.7cm}
\end{table}

\subsection{Policy Effectiveness Validation}\label{sec:valid_robust}

As shown in Fig.~\ref{RL_Train_process}, both the intra-floor and inter-floor policies improve substantially and converge after the first training round. As shown in Table~\ref{tab:diff_round}, the second training round improves SR by only 0.95\% over the first round, indicating that two rounds of training are effective overall but the second round provides limited additional gains. \rev{We also compare alternating RL with joint RL after IL initialization. On HM3D, joint RL achieves 64.7\% SR and 34.25\% SPL, while alternating RL achieves 65.8\% SR and 35.0\% SPL.} To reduce costs, all following experiments use the policies from
the first training round. As shown in
Table~\ref{tab:nav_performance_distill}, both IL and IL+RL (Round 1) achieve
higher SR than VLFM, while IL+RL attains comparable SPL. It indicates that distillation effectively initializes the intra-floor policy for subsequent RL, validating the effectiveness of our factorized policy.

\begin{table}[t]
\centering
\vspace{0.55cm}
\caption{\small{
\rev{Evaluation performance of our method and ASCENT under different test-time perception settings on HM3D.}
}}
\setlength{\tabcolsep}{2.5pt}
\renewcommand{\arraystretch}{0.2}
\scalebox{1.0}{
\begin{tabular}{ccccc}
\toprule
\multicolumn{1}{c}{\textbf{Method}} & \multicolumn{1}{c}{\rev{\textbf{Perception in training}}} & \multicolumn{1}{c}{\rev{\textbf{Perception in testing}}} & \multicolumn{1}{c}{\textbf{SR$\uparrow$}} & \multicolumn{1}{c}{\textbf{SPL$\uparrow$}} \\\\
\midrule
ASCENT & - & G-DINO + YOLOv7  & 60.9 & 29.60 \\\\
ASCENT & - & G-DINO + D-FINE  & 65.4 & 33.50 \\\\
ASCENT & - & GT  & \underline{70.3} & \underline{41.00} \\\\
\midrule
Our & G-DINO + D-FINE & G-DINO + D-FINE & 65.1 & 32.70 \\\\
Our & G-DINO + D-FINE & GT  & 73.9 & 43.43 \\\\
Our & GT & G-DINO + YOLOv7  & 63.8 & 31.76 \\\\
Our & GT & G-DINO + D-FINE  & 65.8 & 35.00 \\\\
Our & GT & GT  & \textbf{80.6} & \textbf{45.26} \\\\
\bottomrule
\end{tabular}
}
\label{tab:general}
\vspace{-0.7cm}
\end{table}

\begin{table}[t]
\centering
\vspace{0.55cm}
\caption{\small{
Evaluation performance of different inter-floor policies under different stair climbing modes on HM3D.
}}
\setlength{\tabcolsep}{4.5pt}
\renewcommand{\arraystretch}{0.3}
\scalebox{1.00}{
\begin{tabular}{ccccc}
\toprule
\multicolumn{1}{c}{\textbf{Inter-floor policy}} & \multicolumn{1}{c}{\textbf{Perception}} & \multicolumn{1}{c}{\textbf{Stair Climbing}} & \multicolumn{1}{c}{\textbf{SR$\uparrow$}} & \multicolumn{1}{c}{\textbf{SPL$\uparrow$}} \\\\
\midrule
Greedy & GT & Waypoint & 80.1 & 44.99 \\\\
Greedy & GT & Teleportation & \underline{86.2} & -- \\\\
\midrule
IL+RL & GT & Waypoint & 79.4 & 44.61 \\\\
IL+RL & GT & Teleportation & \textbf{86.5} & -- \\\\
\bottomrule
\end{tabular}
}
\label{tab:stair_climbing_ablation}
\vspace{-0.7cm}
\end{table}

\begin{table}[t] \centering \vspace{0.55cm} \caption{\small{\rev{ Fine-grained staircase diagnosis on 50 cross-floor episodes under GT perception and waypoint mode. }}} \setlength{\tabcolsep}{1.6pt} \renewcommand{\arraystretch}{0.3} \scalebox{1.05}{ { \begin{tabular}{cccccc} \toprule \multicolumn{1}{c}{\textbf{Metric}} & \multicolumn{1}{c}{\textbf{Approach}} & \multicolumn{1}{c}{\textbf{Traversal}} & \multicolumn{1}{c}{\textbf{Wrong-floor}} & \multicolumn{1}{c}{\textbf{Unnecessary}} & \multicolumn{1}{c}{\textbf{Timing}}\\ \midrule Rate(\%) & 100.0 & 42.0 & 0.0 & 0.0 & 58.0 \\ \bottomrule \end{tabular} } } \label{tab:stair_diagnosis} \vspace{-0.7cm} \end{table}

\begin{table}[t]
\centering
\vspace{0.55cm}
\caption{\small{
Performance improvement of using teleportation  mode on different datasets.
}}
\setlength{\tabcolsep}{4.0pt}
\renewcommand{\arraystretch}{0.3}
\scalebox{1.05}{
\begin{tabular}{cccccc}
\toprule
\multicolumn{1}{c}{\textbf{Method}} & \multicolumn{1}{c}{\textbf{Dataset}} & \multicolumn{1}{c}{\textbf{Perception}} & \multicolumn{1}{c}{\textbf{Stair Climbing}} & \multicolumn{1}{c}{\textbf{SR$\uparrow$}} & \multicolumn{1}{c}{\textbf{SPL$\uparrow$}} \\\\
\midrule
Our & HM3D & D-FINE & Waypoint & 65.8 & 35.0 \\\\
Our & HM3D & D-FINE & Teleportation & \textbf{71.3} & -- \\\\
\midrule
Our & MP3D & D-FINE & Waypoint & 42.6 & 18.2 \\\\
Our & MP3D & D-FINE & Teleportation & \textbf{43.3} & -- \\\\
\bottomrule
\end{tabular}
}
\label{tab:stair_climbing_ref}
\vspace{-0.7cm}
\end{table}

\begin{table}[t]
\centering
\vspace{0.55cm}
\caption{\small{
\rev{Single-floor real-world performance of the wheeled robot under different perception settings.
}
}}
\setlength{\tabcolsep}{8pt}
\renewcommand{\arraystretch}{0.2}
\scalebox{1.05}{{
\begin{tabular}{cccc}
\toprule
\multicolumn{1}{c}{\textbf{Difficulty}} & \multicolumn{1}{c}{\textbf{Trials}} & \multicolumn{1}{c}{\textbf{w/o fine-tuning}} & \multicolumn{1}{c}{\textbf{w/ fine-tuning}} \\\\
\midrule
Easy & 6 & 2/6 & 6/6 \\\\
Medium & 5 & 0/5 & 5/5 \\\\
Hard & 4 & 0/4 & 4/4 \\\\
All & 15 & 2/15 (13.3\%) & 15/15 (100.0\%) \\\\
\bottomrule
\end{tabular}
}}
\label{tab:real_world}
\vspace{-0.7cm}
\end{table}

\subsection{Results and Hypotheses}\label{sec:Phenomena_Hyp}

\rev{As shown in Table~\ref{tab:nav_performance_all}, we compare our method with reported baselines. Under the 0.2m threshold, our method outperforms ApexNav on HM3D by 4.2\% in SR and 1.0\% in SPL, suggesting that introducing multi-floor decision-making improves overall performance. Under the 0.1m threshold, our method outperforms L3MVN* on HM3D by 9.5\% in SR and 10.8\% in SPL,  suggesting that the learned factorized policy with both inter-floor and intra-floor decision-making is more effective than training a single-floor policy alone. Under the 1m threshold, our method exceeds ASCENT by only 0.4\% in SR and 1.5\% in SPL and remains below the reported SOTA end-to-end method XGX on HM3D. Since these results differ in method types, floor settings, learning requirements, and evaluation settings, Table~\ref{tab:nav_performance_all} is used only to motivate the hypothesis that the remaining bottlenecks arise from factors other than the policy, which is further examined through controlled ablations in the following tables.}

As shown in Table~\ref{tab:intrafloor_training_perform}, replacing YOLOv7 with GT perception during intra-floor policy training improves SR by 21.3\% and SPL by 24.6\%, leading us to hypothesize that perception is a bottleneck.

\rev{As shown in Table~\ref{tab:interfloor_ablation_mono}, under GT perception and waypoint-based stair climbing, the learned inter-floor policy and the greedy policy achieve comparable performance. This result suggests that, once the floor-switching decision is reasonable, the remaining performance gap may be more related to the execution of stair traversal than to high-level floor selection. Therefore, we further diagnose staircase-related failures to identify whether the bottleneck comes from approaching the staircase, selecting the target floor, or traversing the staircase.}

\rev{
To identify the main bottlenecks, we randomly sample 100 evaluation
episodes from HM3D v0.1 and manually assign each episode to one
outcome category. For ambiguous cases, the label is determined by
the earliest irreversible error leading to failure. The categories
are defined as follows:
}

$\bullet$ \textbf{Success (A):} The agent actively stops within the success distance threshold and the simulator reports success.

$\bullet$ \textbf{Benchmark (B):} The episode is marked as failure due to potential inaccuracies in the Habitat benchmark evaluation. 

$\bullet$ \textbf{Perception false positive (C):} The agent navigates to an incorrect object caused by the false detection.

$\bullet$ \textbf{Stair climbing (D):} Failures cause repeated exploration on the same floor until the step limit is reached.

$\bullet$ \textbf{PointNav (E):} The agent repeatedly moves around the goal or stops outside the success threshold.

$\bullet$ \textbf{Perception false negative (F):} The agent passes by the goal but fails to detect it.

$\bullet$ \textbf{Policy (G):} The policy leads to exceeding max steps.

$\bullet$ \textbf{Frontier perception (H):} Frontier perception fails to discover valid frontiers, resulting in insufficient exploration.

\rev{As shown in Fig.~\ref{fig:failure_bar}, 61 episodes succeed and 13 episodes are attributed to benchmark errors. Among the remaining 26 valid failures, object-level perception dominates: 10 false positives and 3 false negatives account for 50.0\% of failures, with a 95\% Wilson confidence interval of [32.1\%, 67.9\%]. These perception errors propagate through our modular pipeline because object-goal detections are projected and fused into the semantic frontier map $F^t$. False positives introduce false semantic candidates that may be selected as sub-goals, while false negatives remove true goal candidates from the policy input. Stair climbing is the second largest failure source, with 6 episodes, accounting for 23.1\% of valid non-benchmark failures with a 95\% Wilson confidence interval of [11.0\%, 42.1\%]. Frontier perception accounts for only one episode. These results identify object-level perception reliability as the primary bottleneck and stair-climbing stability as the secondary bottleneck.}

\begin{figure*}[t]
\centering
\includegraphics[scale=0.115]
{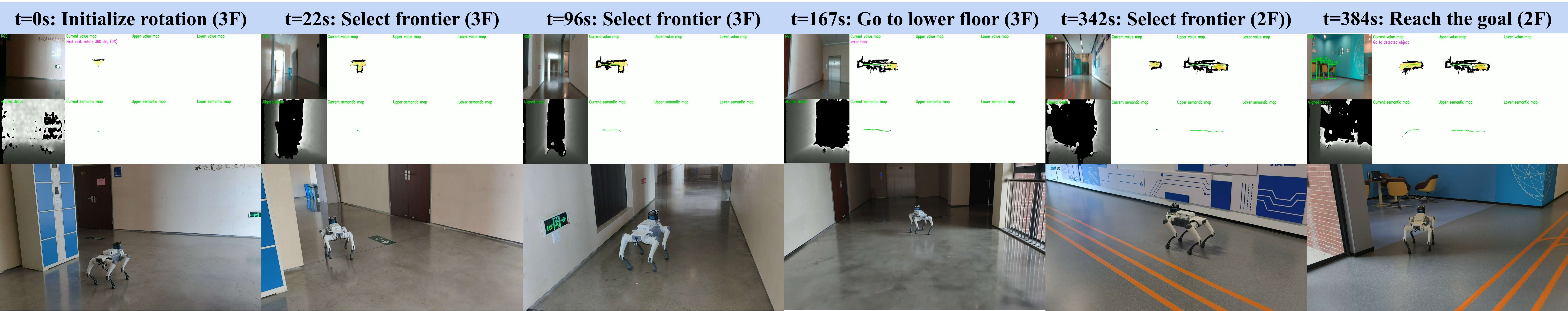}
\caption{\small{\rev{
A real-world multi-floor demonstration using the
legged robot, where the object goal is a chair. After exploring
the initial floor for some time, the robot decides to move to the lower floor, where it successfully locates the goal and completes
the task.
}
}}
\label{fig:all_real_world}
\vspace{-0.8cm}
\end{figure*}

\subsection{Validation of Hypotheses}

As shown in Table~\ref{tab:nav_performance_all_gt}, \rev{under test-time GT perception,} our method improves SR and SPL over ASCENT by 10.3\% and 4.3\%, respectively. It validates our hypothesis that perception failure is one of the root causes of performance bottlenecks. 
Table~\ref{tab:general} compares ASCENT and our method under different \rev{test-time} perception settings. With YOLOv7-based perception, our method improves SR by 2.9\% and SPL by 2.16\% over ASCENT. With ground-truth (GT) perception, the improvements increase to 10.3\% in SR and 4.26\% in SPL. \rev{To isolate the source of the GT gain, we also train the policy with D-FINE-based perception. Using D-FINE for both training and evaluation achieves 65.1\% SR and 32.70\% SPL, comparable to ASCENT under the same setting. Evaluating this D-FINE-trained policy with GT perception raises SR from 65.1\% to 73.9\%, showing that test-time perception errors substantially limit performance. Training with GT perception further increases SR to 80.6\%, indicating that accurate map representations base on GT perception also benefit policy learning. Thus, the GT gain mainly comes from reducing perception-induced map noise during both training and evaluation.}

As shown in Table~\ref{tab:stair_climbing_ablation}, compared with waypoint mode, our inter-floor policy improves SR by 7.1\% in teleportation mode, and greedy policy improves SR by 6.1\% in teleportation mode. In teleportation mode, our policy achieves a 0.3\% SR improvement compared to the greedy policy. 
\rev{Therefore, the learned inter-floor policy is not the main source of performance gain over the greedy rule. Instead, these results indicate that high-level floor switching is already reasonable in most episodes, and the remaining bottleneck mainly lies in low-level stair traversal.} \rev{To further analyze staircase-related failures, we randomly sample 50 multi-floor episodes. As shown in Table~\ref{tab:stair_diagnosis}, staircase approach success reaches 100\%, and both wrong-floor switching rate and unnecessary switching rate are 0\%. However, staircase traversal succeeds in only 42\%, while floor-switch timeout error reaches 58\%. It indicates that the robot reliably reaches the staircase and selects the correct floor but often gets stuck during waypoint-based traversal until timeout. Thus, the bottleneck mainly lies in low-level staircase traversal stability rather than inter-floor decision-making.}

Table~\ref{tab:stair_climbing_ref} reports the improvement from teleportation. It improves SR by 5.5\% on HM3D and 0.7\% on MP3D, showing that better stair-climbing execution benefits multi-floor navigation. As shown in Table~\ref{tab:nav_performance_all}, compared with XGX, the SOTA method on HM3D, our method improves SR and SPL on MP3D by 29\% and 13.1\%, respectively, suggesting better cross-dataset generalization than end-to-end methods.

\subsection{\rev{Real-World Demonstration}}

\rev{We conduct real-world demonstrations with a wheeled robot in an
office and a legged robot in a teaching building. The wheeled
robot is evaluated in 15 single-floor trials across 8 object-goal
categories: person, desk, robotic arm, cabinet, chair, whiteboard,
black box, and white shelf. The trials are classified as easy,
medium, or hard when the goal is initially visible, not visible
but in the same room, or located in a different room. Perception fine-tuning
uses data from the reconstruction of the office, while
evaluation uses separate test episodes. Thus, this evaluation serves
as an environment-adaptation demonstration rather than primary
evidence for the diagnostic conclusion. Table~\ref{tab:real_world} shows that fine-tuning improves the SR from 2/15 (13.3\%) to 15/15 (100.0\%), supporting the perception bottleneck
identified in simulation.
} \rev{We also conduct five demonstrations with the legged robot,
including three multi-floor trials with chair, bicycle, and bin as
goals, and two single-floor trials with chair and bin as goals. In each
multi-floor trial, the robot autonomously explores the current
floor, decides whether to switch floors, and navigates to a
predefined elevator anchor registered as the stair frontier. To
approximate the teleportation mode, floor transitions
are performed by elevator. Only entry and exit are manually
teleoperated, while all other processes are autonomous.
Fig.~\ref{fig:all_real_world} shows a case in which
the robot moves to the lower floor and successfully locates a
chair.
}

\subsection{\rev{Parameter Sensitivity Analysis}}

\begin{table}[t]
\centering
\vspace{0.55cm}
\caption{\small{
\rev{
Parameter sensitivity on HM3D. Default settings: frontier threshold $=1.5$, stair-frontier threshold $=20$, $\Delta k_{\rm intra}=5$, $\Delta k_{\rm inter}=50$, success reward $=2.5$, penalty coefficient $=0.01$.
}
}}
\setlength{\tabcolsep}{15pt}
\renewcommand{\arraystretch}{0.85}
\scalebox{0.95}{{
\begin{tabular}{cccc}
\toprule
\multicolumn{1}{c}{\textbf{Parameter}} 
& \multicolumn{1}{c}{\textbf{Value}} 
& \multicolumn{1}{c}{\textbf{SR$\uparrow$}} 
& \multicolumn{1}{c}{\textbf{SPL$\uparrow$}} \\
\midrule

\textbf{Default (ours)} 
& -- 
& \textbf{65.8} 
& \textbf{35.00} \\
\midrule

\multirow{2}{*}{Frontier thre.} 
& 0.5 & 64.7 & 34.17 \\
& 2.5 & 65.5 & 34.04 \\
\midrule

\multirow{2}{*}{$\Delta k_{\rm intra}$} 
& 2 & 63.4 & 34.79 \\
& 10 & 64.9 & 34.08 \\
\midrule

\multirow{2}{*}{$\Delta k_{\rm inter}$} 
& 25 & 65.6 & 34.96 \\
& 100 & 65.1 & 34.20 \\
\midrule

\multirow{2}{*}{Stair-frontier thre.} 
& 10 & 65.5 & 34.92 \\
& 40 & 65.7 & 34.01 \\
\midrule

\multirow{2}{*}{Success reward} 
& 1 & 64.6 & 33.25 \\
& 5 & 64.2 & 33.94 \\
\midrule

\multirow{2}{*}{Penalty coef.} 
& 0.02 & 65.2 & 34.31 \\
& 0.005 & 65.1 & 31.85 \\

\bottomrule
\end{tabular}
}}
\label{tab:param_sensitivity}
\vspace{-0.7cm}
\end{table}

\rev{Table~\ref{tab:param_sensitivity} reports the sensitivity of key
parameters. The default setting performs best among the tested values. SR varies by only 1.1\% and 0.3\% for the frontier and
stair-frontier thresholds, indicating low sensitivity to
map-generation thresholds. For $\Delta k_{\rm intra}$, overly
frequent or sparse decisions slightly degrade performance, while
$\Delta k_{\rm intra}=5$ balances policy replanning and low-level
execution. A smaller inter-floor interval yields comparable
performance, whereas a larger interval reduces SR and SPL due to
less frequent floor-switching updates. Both the success reward and
step-penalty coefficient affect path efficiency, and an overly small
step penalty noticeably reduces SPL.}

\vspace{-0.2cm}
\section{Conclusions}

We present a diagnostic study of multi-floor ObjectNav using a
modular framework with learnable policies. Our hierarchical
factorization policy includes intra-floor and inter-floor policies,
with a lightweight distilled intra-floor policy providing effective
RL initialization. The diagnostics identify perception and stair
climbing as the primary bottlenecks.

\bibliographystyle{ieeetr}
\end{document}